\documentclass[10pt]{article}
\usepackage[letterpaper]{geometry}
\usepackage{hicss}
\usepackage{times}
\usepackage[none]{hyphenat}
\usepackage{url}
\usepackage{latexsym}
\usepackage{minted}
\usepackage{indentfirst}
\usepackage{graphicx}
\graphicspath{{images/}}
\usepackage[
    style=apa,
  ]{biblatex}
\usepackage{amsmath}
\usepackage{xcolor}
\usepackage{caption}

\title{A Dimensionality-Reduced XAI Framework for Roundabout Crash Severity Insights}

\author{Rohit Chakraborty \\
Ingram School of Engineering \\
Texas State University \\
San Marcos, Texas, USA\\
{\underline{ xuw12@txstate.edu}} \\ \And
Subasish Das, Ph.D. \\
Ingram School of Engineering \\
Texas State University \\
San Marcos, Texas, USA\\
{\underline{ subasish@txstate.edu}} \\
}

\date{}

\begin{document}
\maketitle
\begin{abstract}
Roundabouts reduce severe crashes, yet risk patterns vary by conditions. This study analyzes 2017–2021 Ohio roundabout crashes using a two-step, explainable workflow. Cluster Correspondence Analysis (CCA) identifies co-occurring factors and yields four crash patterns. A tree-based severity model is then interpreted with SHAP to quantify drivers of injury within and across patterns. Results show higher severity when darkness, wet surfaces, and higher posted speeds coincide with fixed-object or angle events, and lower severity in clear, low-speed settings. Pattern-specific explanations highlight mechanisms at entries (fail-to-yield, gap acceptance), within multi-lane circulation (improper maneuvers), and during slow-downs (rear-end). The workflow links pattern discovery with case-level explanations, supporting site screening, countermeasure selection, and audit-ready reporting. The contribution to Information Systems is a practical template for usable XAI in public safety analytics.
\end{abstract}

\subsubsection*{Keywords:}

Explainable AI (XAI), SHAP, Cluster Correspondence Analysis (CCA), interpretable machine learning, roundabout safety, crash severity, decision support systems, transportation analytics.

\section{Introduction}
\vspace{-10pt}

A roundabout is a particular type of at-grade intersection that is widely recognized as a safer alternative to traditional intersections, primarily due to the capacity to reduce conflict points and encourage lower vehicle speeds (\cite{maji_systematic_2025}). Across the United States, roundabouts have been shown to decrease fatalities compared to signalized intersections. Roundabouts have become increasingly important in the United States due to their ability to reduce crash severity, slow down vehicles, and lower conflicts at intersections compared to traditional intersections with traffic signals or stop signs. Even though roundabouts generally improve safety, crashes still happen at these locations, and ongoing research and safety improvements are needed. Studying roundabout safety is necessary because more roundabouts are being built, and traffic conditions continue to change. Roundabouts improve intersection safety mainly by reducing the points where crashes can occur, slowing vehicle speeds, and making decisions easier for drivers. Many studies show fewer serious crashes after roundabouts are built (\cite{polders2015identifying}; \cite{royce2022fix}; \cite{ashqar2024factors}). However, crashes at roundabouts are different from crashes at regular intersections in terms of why they happen, the types of crashes, and who is involved. These crashes contribute to lowering crash severity by removing many of the high-risk crash types. In particular, crashes associated with crossing paths and left-turn movements are eliminated in a roundabout setting. Moreover, the geometric design of roundabouts naturally encourages lower travel speeds, which decreases both the likelihood of a crash and the severity of injuries when crashes do occur (\cite{gross2013safety}). Although, the crash severity might be reduced, however, previous studies identified that driver behavior, design complexity, and unlawful turning can lead to significant number of crashes at roundabouts (\cite{sussman2022turning}). Because of these differences, carefully studying crashes at roundabouts helps identify specific safety problems and ways to fix them.  

The need for continued research on roundabout safety is important because transportation systems keep changing. New technologies are already changing how drivers approach, enter, and circulate in roundabouts. Advanced driver-assistance features, such as Adaptive Cruise Control and Automatic Emergency Braking, can smooth or abruptly alter approach speeds, which may reduce some rear-end risks but also create hesitation at yield lines and shorter accepted gaps (\cite{mehta2023securing}). Lane-keeping and lane-centering can improve lane discipline in multi-lane roundabouts but may also increase late lane changes when markings or signs are confusing \cite{rojas2024automated}. Connected vehicle functions that use roadside units can broadcast geometry and advisory messages (e.g., recommended speed, queue warnings), affecting gap acceptance and speed profiles at entries and exits. Low-speed automated shuttles and highly automated test vehicles interact with human drivers in mixed traffic, sometimes yielding longer than necessary and prompting unexpected stop-and-go behavior. In practice, agencies are also piloting technology-enabled measures at roundabouts, dynamic speed feedback signs, LED-enhanced yield signs, and high-friction surface treatments on approaches, to address wet/low-friction conditions and nighttime visibility. These shifts mean roundabout crash patterns can evolve over time, reinforcing the need for methods that can detect changing combinations of roadway, environment, vehicle type, maneuver, and driver-related factors.

Improving roundabout safety includes both physical changes and educational efforts. Physical changes might involve modifying the roundabout design by adjusting entry angles or lane setups, installing better signs and lights, or adding facilities for pedestrians. Education programs and campaigns to raise awareness among drivers, pedestrians, and cyclists are also important. These efforts work together to reduce both the number and severity of crashes. Therefore, researching roundabout safety, analyzing crashes, and creating effective safety measures remain important. As roundabouts become more common, thorough research and regular safety evaluations are needed. Understanding why crashes happen at roundabouts helps authorities make informed decisions that improve safety and make roundabouts safer for everyone.
46 fatalities occurred at 19 different states across the U.S. between 2005 to 2013, according to a Federal Highway Administration (FHWA) report (\cite{steyn_accelerating_2015}). Another recent study in Michigan showed despite the decrease in fatalities, roundabouts are prone to 58\% more crashes compared to other conventional intersections (\cite{savolainen_evaluating_2023}).  These incidents not only challenge the assumption that roundabouts are inherently safe but also highlight the need for a closer examination of specific crash characteristics at roundabouts. This study will investigate the crashes that occurred at roundabouts in Ohio between 2017 - 2021 to investigate the influencing factors related to such crashes. In Ohio, where roundabouts are increasingly implemented as a safety measure, understanding the nature and causes of these crashes is crucial to enhancing roundabout safety further and informing engineering, enforcement, and educational interventions.

While roundabouts are widely recognized for reducing conflict points and improving intersection safety, existing research has mostly relied on traditional statistical models that examine individual factors in isolation. These methods often fail to capture the complex interactions among vehicle types, roadway conditions, driver behavior, and environmental factors that contribute to crash severity. Moreover, few studies have applied unsupervised learning or interpretable machine learning methods to roundabout crash data, limiting the ability to discover hidden patterns or explain model decisions. This study addresses these gaps by using cluster correspondence analysis (CCA) as a dimension reduction and clustering technique (\cite{velden_cluster_2017}), which enables the grouping of crashes based on shared categorical characteristics in a low-dimensional space. This helps reveal underlying patterns that may not be evident through conventional analysis. In addition, SHapley Additive Explanations (SHAP) is applied to interpret the impact of selected features on crash severity across different clusters (\cite{lundberg_unified_2017}). From a computer science perspective, the study demonstrates how combining unsupervised clustering, dimension reduction, and model explainability provides a transparent, scalable framework for analyzing high-dimensional crash datasets and guiding data-informed safety strategies. Several previous studies have explored explainable AI frameworks to improve traffic safety and determine injury severity level (\cite{amini2022discovering}, \cite{aboulola2024automated}, \cite{khasawneh2025explainable}).  This study also addresses the following research questions:
\begin{itemize}
  \item \textbf{RQ1.} Which combinations of roadway, environmental, vehicle, maneuver, and driver factors tend to co-occur in roundabout crashes?
  \item \textbf{RQ2.} How are these factor combinations associated with injury severity at roundabouts?
  \item \textbf{RQ3.} Within each cluster, which factors most strongly influence severity outcomes, and how do these influences differ across clusters?
\end{itemize}

\section{Methodology}

\subsection{Cluster Correspondence Analysis}
 \vspace{-10pt}
CCA is a robust analytical technique designed to explore relationships within categorical datasets by grouping similar observations (\cite{Rahman2022}). CCA integrates \textit{correspondence analysis} with \textit{K-means clustering}, facilitating both optimal cluster assignment and quantification of categorical variable categories in a reduced-dimensional space (\cite{velden_cluster_2017}). This hybrid approach enhances interpretability by ensuring that category quantifications emphasize differences across clusters.

CCA begins by randomly assigning observations to clusters, establishing an initial $Z_K$ and computing the contingency matrix $F$. Correspondence analysis on $F$ produces a \textit{category quantification matrix} $B$, which is then used to derive the \textit{object coordinate matrix}:

\begin{equation}
Y = \frac{1}{q}(I_n - \frac{1_n 1_n^\prime}{n})ZB
\label{eq:Y_matrix}
\end{equation}

Subsequently, K-means clustering is performed on $Y$ to update $Z_K$, and this process is repeated until convergence is achieved (\cite{velden_cluster_2017}). The final result includes a cluster centroid matrix $G$ and the optimized category quantification matrix $B$.

For visualization, these matrices are rescaled using the factor:

\begin{equation}
\gamma = \left( \frac{K}{Q} \cdot \frac{\text{Tr}(B^\top B)}{\text{Tr}(G^\top G)} \right)^{1/4}
\label{eq:gamma}
\end{equation}

\subsection{SHapley Additive exPlanations}
 
To further interpret the severity outcomes predicted by the machine learning model, this study used SHAP. SHAP provides a consistent and locally accurate method for decomposing a model’s prediction into contributions from each input feature. It is based on the \textit{Shapley value} concept from cooperative game theory, where each feature is treated as a “player” contributing to the final output of the model.

SHAP values represent how much each individual feature contributes to increasing or decreasing the predicted crash severity. A \textit{positive SHAP value} indicates that a feature pushes the prediction toward a more severe outcome, while a \textit{negative value} implies a contribution toward lower severity. The \textit{magnitude} of the SHAP value reflects the strength of the feature’s influence on the model's output. For a given model $f(x)$, the SHAP value $\phi_i$ for a feature $x_i$ is computed using the following formulation (\cite{lundberg_unified_2017}):

\begin{equation}
f(x) = \sum_{S \subseteq F \setminus \{i\}} \frac{|S|!(|F| - |S| - 1)!}{|F|!} \left[ f(S \cup \{i\}) - f(S) \right]
\label{eq:shap}
\end{equation}

\subsection{Data Description and Variable Selection}
\begin{figure*}[!t]
  \captionsetup{labelfont={color=black},textfont={color=black}}
  \centering
  \includegraphics[width=0.95\textwidth]{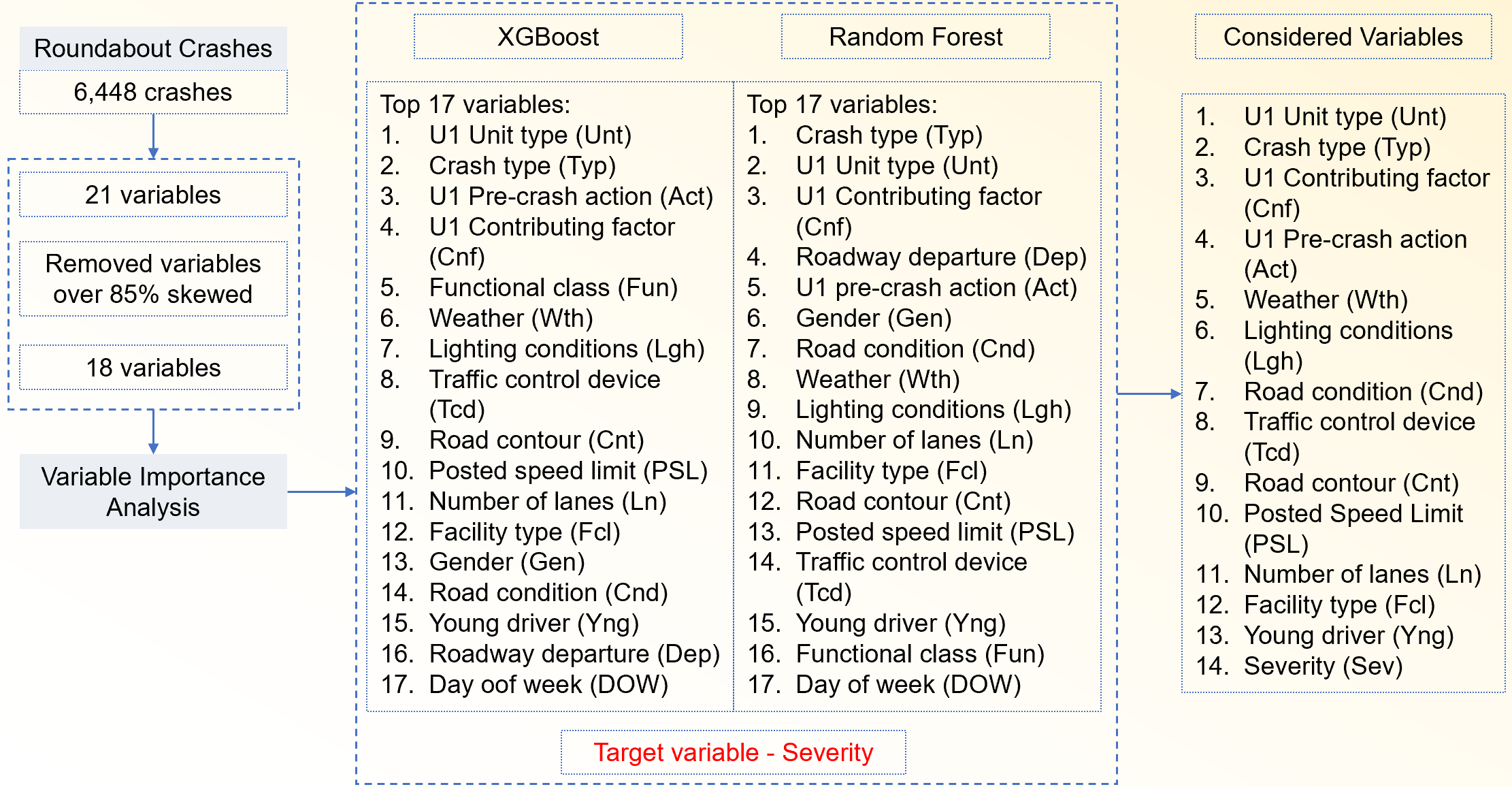}
  \caption{List of Considered Variables}
  \label{fig:considered variables}
\end{figure*}

\vspace{-10pt}
The dataset initially consisted of 6,448 roundabout-related crashes in Ohio between 2017 and 2021, with 21 categorical variables describing vehicle characteristics, roadway conditions, environmental factors, and driver behavior. To identify the most influential variables contributing to crash severity at roundabouts, a systematic variable selection process was conducted, as illustrated in Figure \ref{fig:considered variables}. To ensure meaningful and unbiased analysis, variables with more than 85\% skewed distribution (i.e., dominated by a single category) were removed. This preprocessing step resulted in a refined set of 18 variables. These variables were then subjected to a Variable Importance Analysis using two machine learning algorithms, XGBoost (\cite{chen_xgboost_2016}) and Random Forest (\cite{breiman_randomforest_2002}), both of which are known for their robustness in handling high-dimensional and categorical data. The two different feature screening methods relied on two tree-ensemble measures. Random Forest importance was the mean decrease in Gini impurity across all splits and trees. XGBoost importance was the cumulative gain in the objective (loss) from all splits on a variable across boosted trees. Importance were computed on cross-validated training folds, normalized and averaged, and variables with consistently above-median scores were retained.

\begin{figure*}[htbp]
    \centering

    \includegraphics[width=0.49\linewidth]{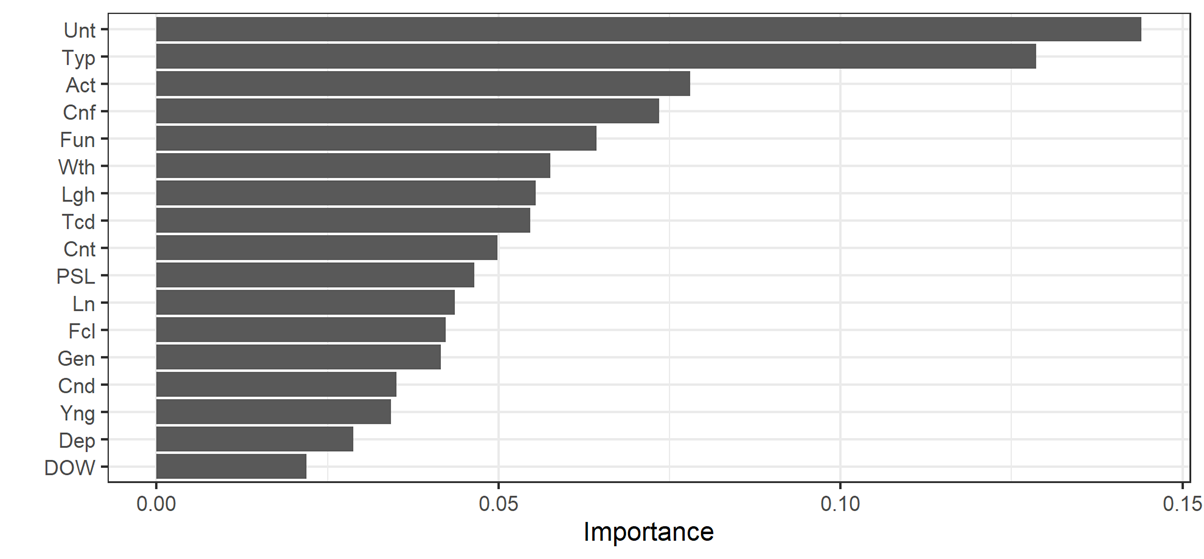}
    \hfill
    \includegraphics[width=0.49\linewidth]{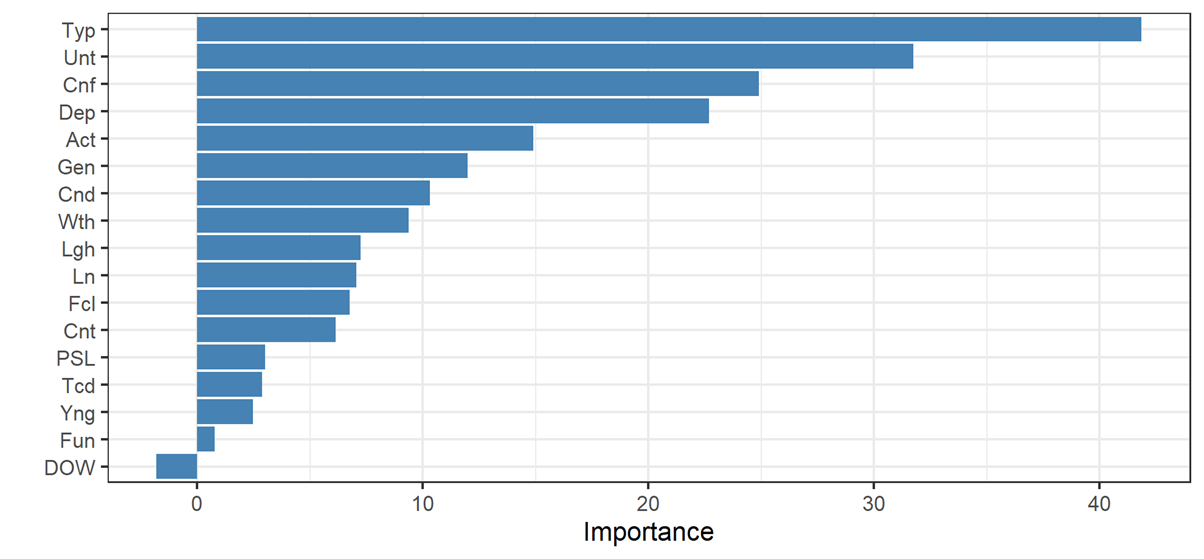}

    \par\smallskip
    \makebox[\linewidth]{(a) XGBoost \hspace{0.40\linewidth} (b) Random Forest}
    \caption{Variable Importance Plots}
    \label{fig:VIP}
\end{figure*}

Based on the consensus across both models, a final set of 14 variables was selected for CCA. These included vehicle type, crash type, contributing factor, pre-crash action, and roadway and environmental conditions (e.g., weather, lighting, road contour, posted speed limit, number of lanes, and facility type). The variable “Severity” was retained as a reference outcome to support interpretability and discussion of CCA-based crash patterns. This data-driven approach ensured that only relevant and non-redundant variables were used in the clustering analysis, leading to more reliable pattern detection and interpretation in the context of roundabout crashes. The plots in Figure \ref{fig:VIP}(a) and Figure \ref{fig:VIP}(b) show the variable importance analysis using XGBoost and Random Forest models respectively. Both models selected and highlighted the importance score of each variable on crash severity.

\section{Results}
\vspace{-10pt}
This section presents the key findings derived from the roundabout crash dataset using a multi-step analysis pipeline. Based on a refined set of variables was selected for CCA, which was applied to discover distinct crash patterns across the dataset. The resulting clusters were then interpreted both contextually and visually. SHAP analysis was used to explain how individual features contributed to severity predictions within each cluster, providing model-level transparency and deeper insight into crash dynamics. 

\subsection{Cluster Correspondence Analysis}
 
This section includes a description of the findings from this study. Table \ref{tab:roundabout_clusters} provides a summary of the centroids, variability, and sizes of the clusters identified from the roundabout crash data using CCA. The centroids (Dim 1 and Dim 2) indicate the central positions of each cluster in the data space, while the within-cluster sum of squares reflects the degree of variability among crashes within each cluster. The size column indicates the number of crashes assigned to each cluster, with Cluster 1 containing the most crashes (2,453) and Cluster 4 the fewest (1,298). The table illustrates how crashes were grouped based on shared features and the relative internal consistency of each cluster.
\begin{table}[htbp]
\centering
\caption{Centroids and size of the clusters}
\resizebox{\columnwidth}{!}{%
\begin{tabular}{lrrrr}
\hline
\textbf{Cluster} & \textbf{Dim 1} & \textbf{Dim 2} & \textbf{Within Cluster Sum of Squares} & \textbf{Size} \\
\hline
Cluster 1 & -0.0096 & -0.0053 & 0.0424 & 2453 \\
Cluster 2 & -0.0026 &  0.0036 & 0.0327 & 1396 \\
Cluster 3 &  0.0048 &  0.0165 & 0.0396 & 1301 \\
Cluster 4 &  0.0161 & -0.0103 & 0.0408 & 1298 \\
\hline
\end{tabular}%
}
\label{tab:roundabout_clusters}
\end{table}

\begin{figure*}[htp]
    \centering
    \includegraphics[width=1\linewidth]{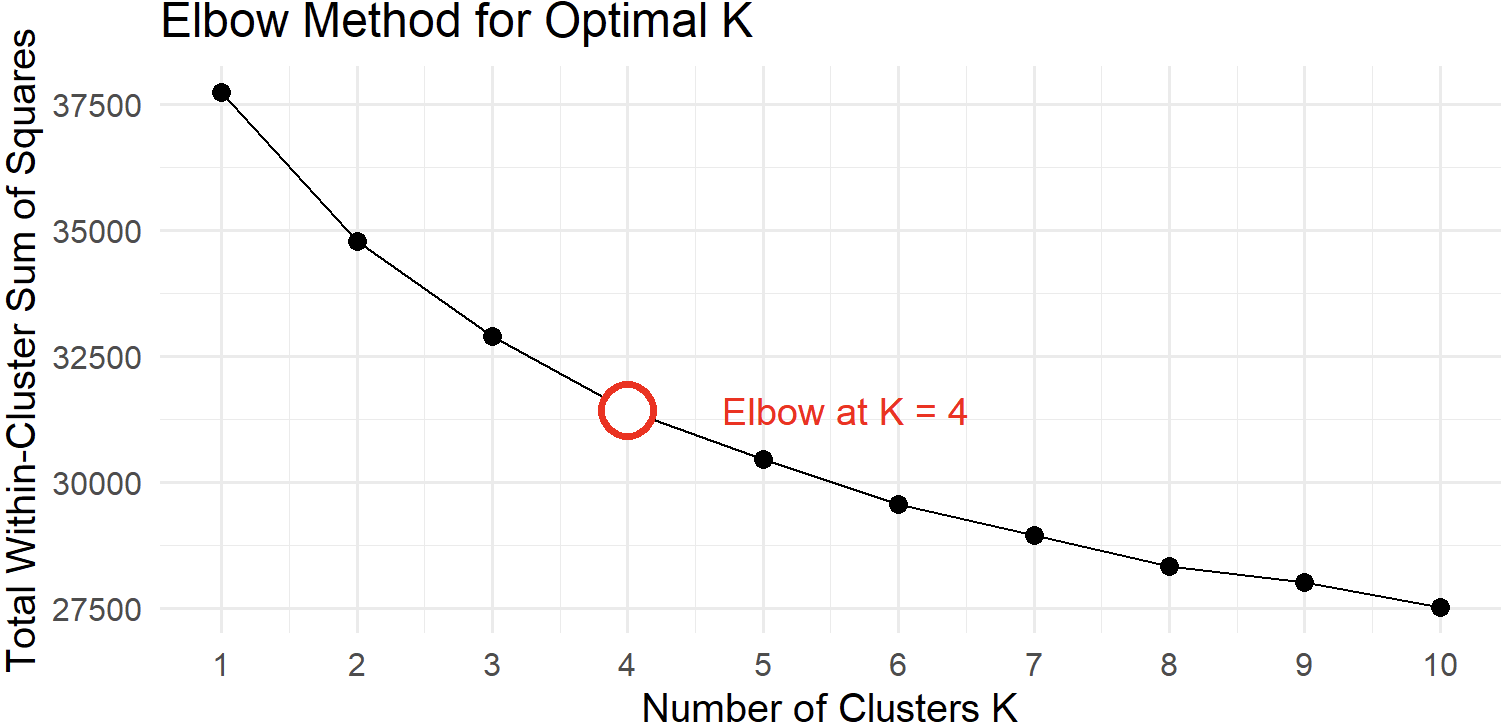}
    \caption{Elbow Method for Optimum Number of Cluster Identification}
    \label{fig:Elbow}
\end{figure*}

Figure \ref{fig:Elbow} plots the total within-cluster sum of squares (WCSS) for \(k\)-means applied to the CCA coordinates across \(K=1\)--\(10\). WCSS measures within-cluster dispersion; lower values indicate tighter, more coherent clusters. The curve drops steeply from \(K=1\) to \(K=3\) and shows a clear change in slope at \(K=4\) (red circle). Beyond \(K=4\) the reductions are incremental and nearly linear, indicating diminishing returns from adding clusters. This pattern supports selecting four clusters: \(K=4\) captures the dominant structure in the data, whereas larger \(K\) mainly split existing groups without revealing new structure. The four-cluster solution also yields balanced sizes and interpretable roundabout crash patterns, so \(K=4\) is retained.

\begin{figure*}[htbp]
    \centering

    \includegraphics[width=0.48\linewidth]{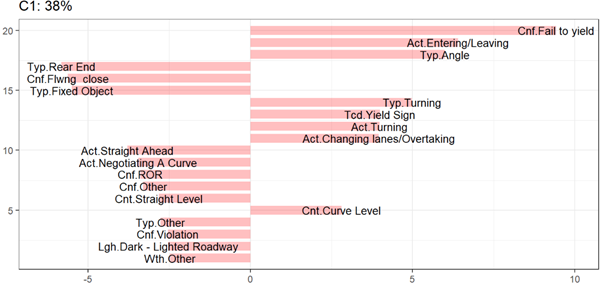}
    \hfill
    \includegraphics[width=0.48\linewidth]{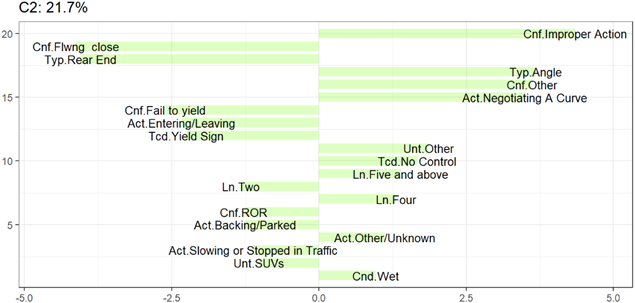}

    \par\smallskip
    \makebox[\linewidth]{(a) Cluster 1 \hspace{0.40\linewidth} (b) Cluster 2}

    \par\bigskip

    \includegraphics[width=0.48\linewidth]{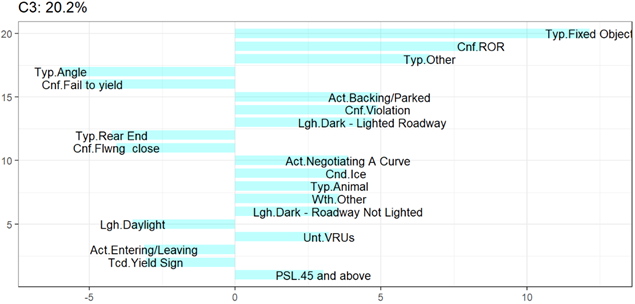}
    \hfill
    \includegraphics[width=0.48\linewidth]{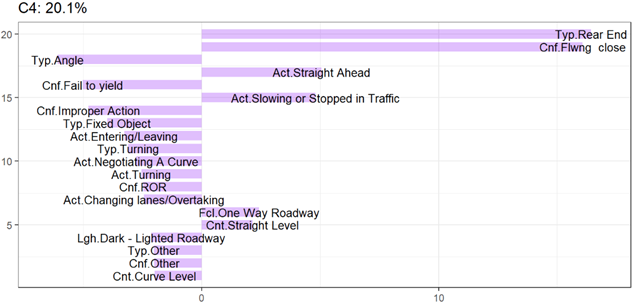}

    \par\smallskip
    \makebox[\linewidth]{(c) Cluster 3 \hspace{0.40\linewidth} (d) Cluster 4}

    \caption{Clusters on Roundabout Crashes}
    \label{fig:clusters}
\end{figure*}

\subsubsection{Cluster 1 (C1) – Entry and Yield-Related Conflicts at Roundabouts:} 

Cluster 1 in Figure \ref{fig:clusters}(a), accounts for 38\% of all roundabout crashes in the dataset and primarily highlights crashes related to improper entries and yielding behavior. The most prominent factors in this cluster are “Fail to Yield”, “Entering/Leaving”, and “Angle”-type crashes, all of which are closely tied to the roundabout’s unique structure that requires drivers to yield before entering and maintain awareness of circulating traffic. In this cluster, crashes are often triggered by vehicles entering the roundabout without properly yielding to circulating traffic. These crashes typically result in angle collisions, which are common when vehicles merge at improper times or misjudge the speed or gap of other road users already in the roundabout (\cite{montella2011identifying}). The strong presence of the “Yield Sign” traffic control variable in this cluster supports this interpretation, indicating that even when yield signs are present, some drivers either fail to notice them or choose not to follow the right-of-way rule.

\subsubsection{Cluster 2 (C2)- Improper Maneuvers and Driver Misjudgment Crashes at Roundabouts: }

Cluster 2 in Figure \ref{fig:clusters}(b) includes 21.7\% of all roundabout crashes and is largely shaped by driver behavior such as “Improper Action”. The cluster also includes “Angle”-type crashes, and “Negotiating a Curve”, suggesting that these crashes may be associated with confusion or sudden movement while adjusting to the roundabout’s curvature. This is particularly the case in wet conditions, where stopping distance increases, and minor misjudgments can lead to collisions. The inclusion of “Wet” roadway conditions supports this interpretation, as even a slight loss of traction may turn a following-distance mistake into a crash (\cite{saccomanno2008comparing}). The presence of “Five and above” and “Four” lane roads in this cluster may also contribute, as multi-lane roundabouts require greater awareness of both lane positioning and the behavior of vehicles in adjacent lanes, adding to the complexity.

\subsubsection{Cluster 3 (C3)- Fixed Object and Environmental Hazard Crashes at Roundabouts:}

Cluster 3 is highlighted in Figure \ref{fig:clusters}(c). It represents 20.2\% of the total roundabout crashes, is primarily characterized by crashes involving vehicles striking fixed objects, especially under conditions of poor visibility or environmental hazards. The most prominent variable in this cluster is “Fixed Object” crash type, strongly linked with “Run-Off-Road (ROR)” contributing factors, suggesting that drivers are either losing control of their vehicles or misjudging the roundabout’s curvature and geometry, causing them to veer off the intended path. Previous studies indicated that higher approach speed and higher traffic volumes can also contribute to ROR crashes at roundabouts (\cite{burdett2017evaluation}). The environmental conditions in this cluster further support this pattern. Crashes occurring during “Dark – Lighted Roadway” and “Dark – Not Lighted Roadway” conditions point toward limited visibility, which can affect a driver’s ability to judge entry and exit points, lane positions, or the location of medians, signs, or other infrastructure within the roundabout. Such situations are worsened when combined with ice, animal presence, or other unexpected hazards on the road. These environmental risks can disrupt a driver’s attention and reaction time, making it difficult to maintain control on a circular path. Another notable feature of this cluster is the involvement of backing or parked vehicle actions, which are generally unusual at roundabouts. This may indicate confusion or misbehavior, such as illegal maneuvers or incorrect vehicle positioning, especially in larger roundabouts or at roundabout-adjacent areas like splitter islands or exits.

\subsubsection{Cluster 4 (C4)- Rear-End and Following-Distance Crashes in Circulating Roundabout Flow:}

Cluster 4 in Figure \ref{fig:clusters}(d), represents 20.1\% of all roundabout crashes and is strongly associated with rear-end collisions caused by close following distances, particularly during routine or slowed traffic flow within the roundabout. The most influential variables in this cluster are “Rear End” crash type and “Following Too Closely” as a contributing factor, clearly pointing to situations where vehicles collide because drivers are not maintaining sufficient space from the vehicle ahead. This type of crash is common in roundabouts, where stop-and-go movement is frequent. Vehicles may slow down unexpectedly when entering or yielding, or during lane changes while exiting. If a driver is distracted or assumes traffic will keep flowing without interruption, they may fail to react in time. This pattern is emphasized by the variable “Slowing or Stopped in Traffic,” which suggests that a significant number of these crashes occur when vehicles are reducing speed or are momentarily halted, situations that require constant attention from the driver behind. Previous studies also identified that rear-end crashes are frequent at roundabouts, especially if the vehicles are following the lead vehicle too closely (\cite{burdett2016analysis}).

\subsection{SHAP Interpretation}
 \vspace{-10pt}
SHAP values help explain how each feature in a machine learning model influences the predicted outcome, in this study, the severity level of roundabout crashes. The beeswarm SHAP plots display individual crash instances along the horizontal axis based on the SHAP value for each feature. A positive SHAP value indicates that the feature contributes to a higher severity prediction, while a negative value suggests the feature lowers the severity. Each dot represents a crash case and is color-coded by the predicted crash severity level: red for killed or severe injury (KA), blue for moderate or minor injury (BC), and green for no injury (O) crashes. Features are listed vertically by their overall impact on predictions in that cluster, allowing to identify which variables are most important in influencing crash outcomes. 

\begin{figure*}[!t]
  \centering
  \includegraphics[width=0.95\textwidth]{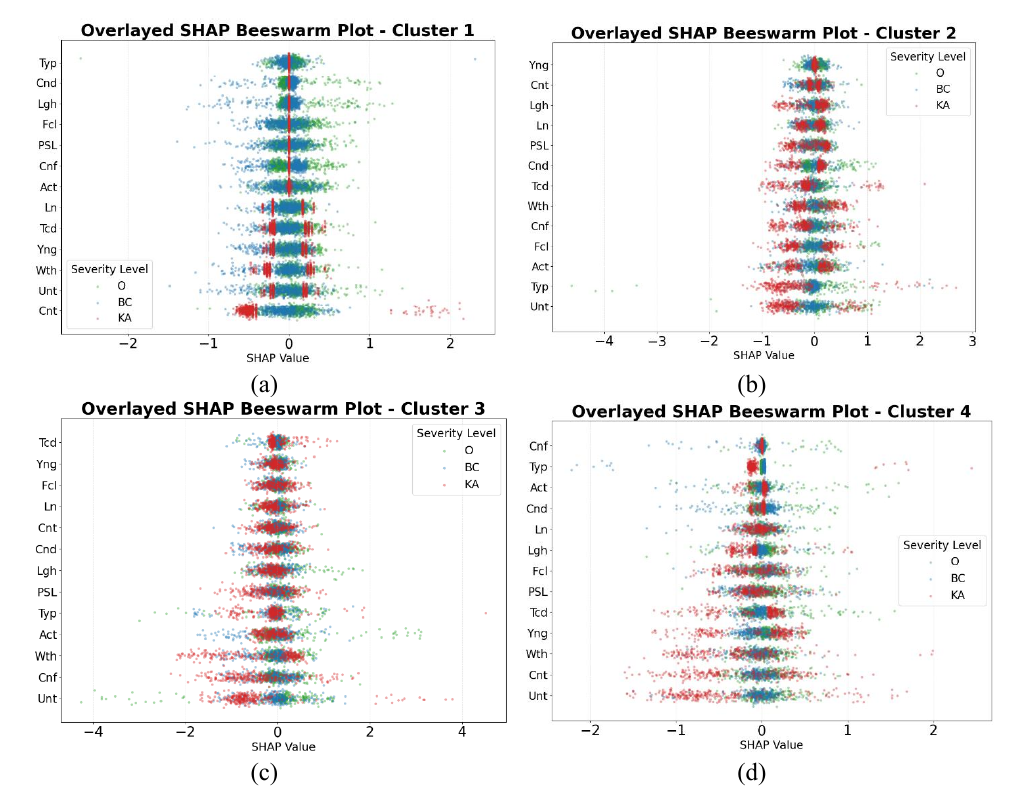}
  \caption{SHAP Value Plots showing Feature Importance for Crash Severity by Clusters.}
  \label{fig:shap_2x2_manualsubcaption}
\end{figure*}

\subsubsection{Cluster 1:}

Cluster 1 includes crashes that typically occur during entry into roundabouts, often due to failure to yield or merging errors. The SHAP analysis in Figure \ref{fig:shap_2x2_manualsubcaption}(a) shows which features most influence the model’s prediction of crash severity for this cluster. SHAP analysis identified unit type as one of the important features. Crashes involving larger vehicles, such as trucks or buses, tend to contribute to higher severity outcomes, likely due to their mass and limited maneuverability during merging. Contributing factor was also found to be important, particularly when failure to yield was involved. These cases often lead to angle-type crashes, which the SHAP values suggest are more likely to result in serious injuries. Crash type and road contour also influenced severity predictions. 

\subsubsection{Cluster 2:}

Cluster 2 includes crashes linked to improper driving actions and misjudgment inside the roundabout, especially under wet or multi-lane conditions. The SHAP analysis in Figure \ref{fig:shap_2x2_manualsubcaption}(b) shows how various features affect the model’s prediction of crash severity for this group. SHAP analysis identified unit type as one of the important features. Crashes involving certain vehicle types, particularly those with reduced control or visibility, are associated with higher severity outcomes in this cluster. Driver action was also found to influence severity. Improper maneuvers such as abrupt lane changes or turning across lanes are common in multi-lane roundabouts and can lead to more serious crashes, which is reflected in the positive SHAP values for KA outcomes. Weather conditions appeared as an important environmental factor. Crashes occurring during rain or on wet surfaces are associated with increased severity, likely due to reduced tire traction and longer stopping distances.

\subsubsection{Cluster 3:}

Cluster 3 includes crashes that mostly involve vehicles hitting fixed objects or losing control under challenging environmental conditions, such as poor lighting or icy surfaces. The SHAP analysis in Figure \ref{fig:shap_2x2_manualsubcaption}(c) highlights how different features contributed to the model’s prediction of crash severity in this group. Severe outcomes are often linked with certain vehicle types that may be more difficult to control under slippery or poorly lit conditions. The positive SHAP values for red points indicate that these vehicles increase the likelihood of more serious crashes. Contributing factor also played an important role in this cluster. Cases involving roadway departure or loss of control showed a strong influence on severity predictions. These behaviors are consistent with fixed object crashes where drivers veer off the roadway, often due to distraction or low visibility. SHAP analysis further identified weather and lighting condition as important features. Crashes in dark or icy environments are more likely to be severe, as shown by the concentration of red SHAP values. These conditions reduce reaction time and visibility, making fixed-object impacts more dangerous.

\subsubsection{Cluster 4:}

Cluster 4 reflects crashes that mainly involve rear-end collisions due to close following distances in circulating roundabout traffic. These crashes typically occur during routine flow where vehicles slow down unexpectedly. SHAP analysis in Figure \ref{fig:shap_2x2_manualsubcaption}(d) identified road contour and weather condition as influential features. Crashes occurring on level road segments and during adverse weather were more likely to result in higher severity, possibly due to reduced traction and delayed braking. Contributing factor played a strong role in shaping severity. Many crashes in this cluster were associated with following too closely, which increases the likelihood of rear-end collisions when traffic suddenly slows. SHAP values show that such behaviors consistently push the model toward predicting more severe outcomes. Other features, such as crash type and driver action contributed meaningfully as well. The presence of “rear-end” as a crash type and “slowing or stopped in traffic” as an action often corresponded with higher SHAP values, reinforcing the pattern that inattentiveness or aggressive driving in circulating flow is a key severity driver.

\section{Discussions}

The findings showed that roundabout crashes group into a small number of consistent patterns, and that injury severity links to specific combinations of lighting, surface condition, speed environment, crash type, and maneuvers. For public agencies, this supports practical actions such as screening sites by pattern, aligning countermeasures to the dominant risks (e.g., yield control and lane guidance at entries; surface friction and drainage on wet approaches; lighting upgrades where darkness and fixed-object crashes co-occur), and evaluating outcomes with the same pattern logic. From an Information Systems perspective, the study illustrated how an interpretable analytics pipeline (CCA for pattern discovery followed by a supervised model with SHAP for explanation) can be embedded in decision tools. Two design lessons stand out: (i) pair pattern-level views for network screening with case-level explanations for site diagnosis, and (ii) present support counts and uncertainty alongside importance ranks to maintain appropriate trust. The approach is suitable for dashboards, annual safety reports, and after-action reviews.

\section{Conclusions}

This study presented a data-driven approach to evaluate roundabout crash patterns and severity outcomes using techniques rooted in machine learning and interpretable data analysis. By applying CCA as a dimension reduction and unsupervised clustering method, the study was able to uncover latent groupings in categorical crash data. These groupings revealed meaningful distinctions between different crash scenarios, such as failure to yield at entry, improper maneuvers, environmental hazards, and following-distance errors, across four unique clusters. To further understand what influences crash severity within each cluster, SHAP was used to interpret the relative importance of individual features in the crash severity prediction model.

The findings showed that entry-related crashes involving large vehicles and failure to yield were particularly associated with higher severity outcomes. Crashes due to improper actions inside multi-lane roundabouts, especially during wet conditions, were also found to be risky. Environmental factors such as poor lighting, high speeds, and loss of control played a central role in fixed-object crashes. Rear-end collisions from following too closely in circulating traffic were frequent and often worsened by inattentiveness and abrupt slowdowns. These insights offer practical implications for both traffic engineers and policymakers. Enhancing signage and geometric design at entry points, enforcing lane discipline, installing better lighting, and promoting safe driving distances within roundabouts are among the recommended strategies. This study directly addressed the three research questions. For RQ1, CCA revealed four clear patterns of co-occurring factors at roundabouts, including entry/yield conflicts, multi-lane maneuver issues in wet conditions, run-off-road/fixed-object events in dark settings with higher posted speed limits, and circulating rear-end conflicts. For RQ2, a supervised severity model with SHAP linked these patterns to injury outcomes, showing that combinations involving darkness, higher speed limits, and fixed-object crashes were associated with higher severity, while clear weather and lower speeds were associated with lower severity. For RQ3, within-cluster SHAP results identified the strongest severity drivers in each pattern, for example, fail-to-yield and angle type at entries; improper maneuver under wet, multi-lane conditions; lighting and fixed-object crashes at night; and following-too-closely in circulating flow. Together, these findings connect specific factor combinations to severity pathways and point to practical steps, improved yield control and lane guidance at entries, surface friction and drainage upgrades, better lighting, and approach speed/queue warnings, to reduce harm at roundabouts.

From a methodological standpoint, this study highlights how interpretable machine learning techniques can improve traffic safety research by not only identifying patterns but also explaining the logic behind predictive models. The use of CCA to reduce dimensionality and reveal structure in high-cardinality categorical data, followed by SHAP to provide granular explanations, offers a useful template for similar applications in transportation and other public safety domains.

This study also contributes to Information Systems by showing a practical path to transparent, auditable decision support for public-sector safety. The CCA and SHAP workflow connects unsupervised pattern discovery with supervised, explainable severity modeling, producing outputs that non-experts can read and act on. It demonstrates how model logic can be surfaced to end users (what factors matter, under which conditions, and why), how those explanations align with domain mechanisms, and how they can drive concrete choices about countermeasures and program management.

This study also acknowledged some limitations. The CCA technique, while effective at uncovering broad patterns, does not incorporate temporal dynamics, such as how crash characteristics evolve over time. Also, the SHAP interpretations are based on model-trained associations and do not necessarily capture causal relationships. Moreover, the current study did not incorporate real-time or vehicle trajectory data, which could enhance understanding of driver behavior leading up to crashes. Geographical limitations is also acknowledged as this study was focused on only Ohio data. Future research can address these limitations by integrating temporal sequence modeling, multi-modal sensor or telematics data, and advanced explainability techniques for time-series or graph-based crash representations. Expanding the framework to include spatial mapping and simulation environments may also help support deployment-level decision-making for safer roundabout design.

\printbibliography

\end{document}